# Industrial Machine Tool Component Surface Defect Dataset

Tobias Schlagenhauf[1], Magnus Landwehr[1] and Jürgen Fleischer[1]

[1]Karlsruhe Institute of Technology

*Abstract*— Using machine learning (ML) techniques in general and deep learning techniques in specific needs a certain amount of data often not available in large quantities in technical domains. The manual inspection of machine tool components and the manual end-of-line check of products are labor- intensive tasks in industrial applications that companies often want to automate. To automate classification processes and develop reliable and robust machine learning-based classification and wear prognostics models, one needs real-world datasets to train and test the models. The dataset is available under https://doi.org/10.5445/IR/1000129520.

## I. Introduction & Background

The classification of real-world objects in an industrial context has been gaining attention since the dawn of powerful deep learning-based models as proposed for instance by (LeCun, Bengio, Hinton 2015). Contests like the ImageNet Contest (Russakovsky et al. 2014) have driven the development of deep learning architectures for large-scale data applications. The use of robust deep learning-based models in an industrial context becomes a critical aspect concerning autonomous machines and production facilities. To reliably use an intelligent algorithm for the automatic end-of-line check of products or the inspection of machines and machine components to realize predictive maintenance, the model must be trained on data representing the underlying process and data distribution. Although until now, there are some applications of deep learning in more technical domains like the classification of malignant tumor cells in images (Tang et al. 2019), the long tail of possible technical uses still lacks data to cover most of the more technical cases in the fields of life science, medicine, or industry and production. One important subfield for classification in an industrial context is the inspection of surfaces, be it on rails (Faghih-Roohi et al. 2016), on concrete (Koch et al. 2015), on wood (He et al. 2019a), or metallic surfaces of products (He et al. 2019b) or machine tool components (Schlagenhauf, Ruppelt, Fleischer 2020). In all these cases, the goal is to detect anomalies on an otherwise intact surface to evaluate the condition of a certain part. Especially metallic surfaces that are found on most machine tool components like ball screw drives (BSD), roller bearings, or linear rail guides are important machine tool components for industrial applications. Robust models for the detection of failures on a wide range of metallic surfaces have the potential to reduce high costs because unforeseen machine breakdowns are prevented and costs for the tedious manual inspection of machine tool components are saved. In a previous work (Schlagenhauf, Ruppelt, Fleischer 2020), the authors published an approach to classify defects on BSD. The model presented there was trained on a dataset like the one presented here. The same sensor system was used in both cases. To the best of our knowledge, there has been, up to now, no detailed description or publication of a real-world BSD surface defect dataset. The dataset has been generated under a realistic setup and consists, amongst others, of real-world examples that are difficult to classify, even for domain experts. The authors provide a labeled dataset in a domain that needs substantial domain knowledge to label the data. Secondly, the dataset has a small interclass variance and a large intraclass variance together with defects in different sizes which makes correct classification a task that is not easy to solve at all. Another unique characteristic of the dataset is that it contains the progression of failures which makes it a ready-to-use dataset for wear prognosis. Based on this, thirdly, the dataset provides the ML community with a dataset to train and test ML models and can be used as a kind of benchmark dataset for defect classification and prognostics models in the industrial context.

## II. Experimental Setup

### A. Ball Screw Drives

BSD are among the main machine tool components used for the translation of rotary motion into linear motion to move, for instance, the axes of machine tools. Here, the BSD is one the most wear-prone machine tools (Haberkern 1998) and one of the machine tool components mainly responsible for the unplanned breakdown of machine tools (Fleischer et al. 2009). In addition to being used in the machine tool industry, BSD are widely used components in industries like the power generation industry, the medicine industry, the automation industry, and the robotics industry. Since BSD are tribological systems (Sommer, Heinz, Schöfer 2018), wear is one of the tribological elements. Based on the classical bathtub curve (Heise 2002) and according to (Haberkern 1998), (Schopp 2009), and (Spohrer 2019), the BSD fails because of one of three failure modes (Figure 1).

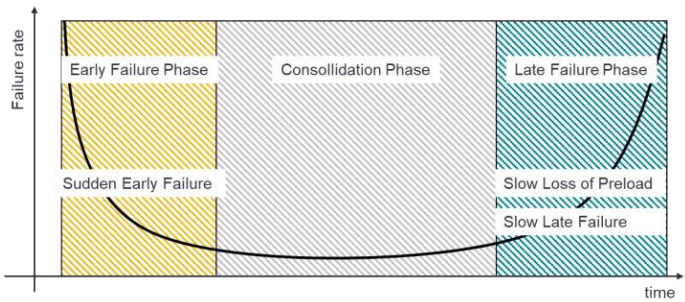

*Figure 1: Failure modes of a BSD based on the bathtub curve*

The first mode is the so-called *sudden early failure* in the early failure phase just after start of the operation. Sudden early failure results e.g. from production failures or because of an incorrect installation of the component. This leads to enormous wear or an instant jam of the system which in consequence leads to the early breakdown of the system.

In the second stage, the bathtub curve shows a so-called *consolidation phase* in which normally no failures occur. This is because the system is run in and there is no longer any reason caused by the system for a sudden failure. In this phase, natural wear of the system takes place, which in the case of the BSD mainly consists in *surface disruption*, *abrasion,* and *adhesion*.

- *Surface disruption* expresses itself in pitting on the surface of the system which then, in turn, accelerates the other two wear mechanisms.
- *Abrasion*, is due to small particles in the system which either come from the system itself or are particles like metallic chips or pollution introduced from the outside.
- *Adhesion* is due to adhesive effects, which leads to larger surface breakouts. These breakouts are then prevalent in the system and lead to acceleration of the other wear mechanisms.

If the wear progresses, the BSD fails either due to the so-called *slow loss of preload* (second failure mode), which is in principle a phenomenon of clearance such that the system can no longer fulfill the proposed task. The third failure mode is the so-called *slow late failure*. This failure mode is mainly due to surface disruptions, which impedes further operation of the system. Since all failure mechanisms affect each other, the true failure mode depends on the type of process and on the system requirements. In this work, the focus is laid on the failure mechanism of the surface disruption and the resulting pitting since the latter can be observed in image data. It is known that the pitting on the surface of the spindle starts at some point in time as single small failures on the surface which then grow in size and finally lead to the failure of the system. Figure 2 shows a heavily worn part of a spindle conceivable in practice.

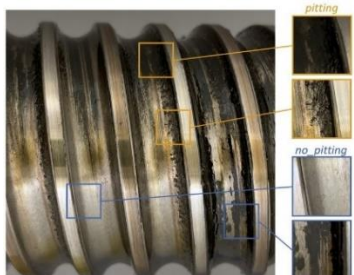

*Figure 2: Example of a worn BSD spindle*

The here generated dataset is based on a sequence of such images taken during a destruction test at the wbk Institute of Production Science at Karlsruhe Institute of Technology. The sensor setup is explained together with the test bench in the next section.

### B. Sensor System

The sensor system used for creation of the image dataset is depicted in Figure 3.

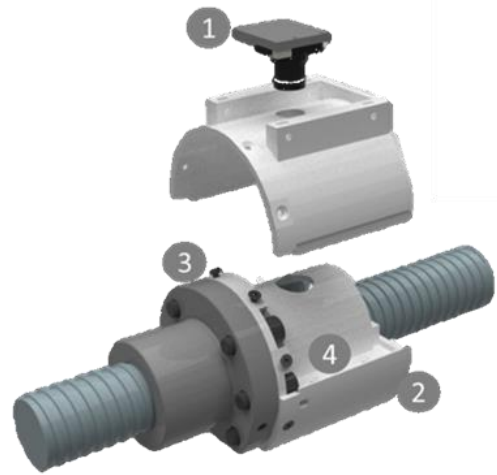

*Figure 3: Sensor system used for image generation*

The system is mounted onto the nut of the BSD using a mounting adapter numbered with #3. The camera (#1) looks through a hole in the so-called diffusor (#4) onto the spindle. Since turning the spindle leads to a linear motion of the nut and the spindle is turning underneath, the camera gets to see all raceways of the spindle. Using this setup, the whole spindle can be photographed. #2 is a manufactured housing enclosing the spindle which is used to ensure uniform lighting conditions during the experiment. Additionally, the housing protects the camera from pollution. An important part of the system which is responsible for lightning of the images is the so-called diffusor which also implements the light sources. The light sources are two standards LED stripes mounted onto the surface where #4 is located. The diffusor itself is 3D- printed and consists of a semitransparent plastic leading to diffuse light. Since the LEDs are not pointed onto the spindle but directly onto the housing, the light does not get directly onto the spindle but is reflected by the housing and then further made more diffuse bypassing the diffusor. During tests, this setup was found to be yielding the best results for our purpose. The used camera system is a standard Raspberry Pi V2 microcontroller camera which is a good tradeoff between resolution, costs, and necessary mounting space. The camera is set up to take images with a resolution of 2592x1944 pixels per image.

### C. Test Setup

The dataset is generated on a test bench located at the Institute of Production Science at the Karlsruhe Institute of Technology. The test bench is depicted together with the mounted camera systems in Figure 4.

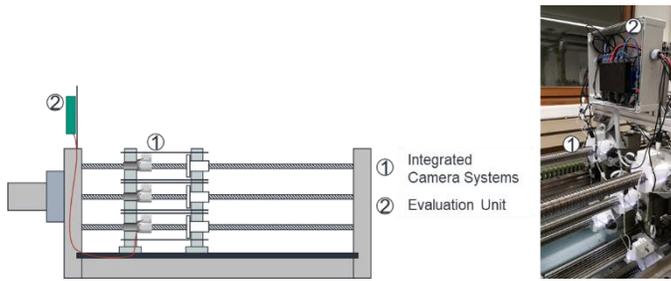

*Figure 4: Test bench with mounted camera systems for image generation*

The test bench is constructed such that a maximum of five spindles can be worn in parallel. The spindles are positioned like the five on a dice, with the middle spindle being the leading spindle connected to the motor. The other four spindles are operated by a chain drive connected to the central spindle, thus it is ensured that all spindles are operated in the same way. The spindles used are standard 32mm diameter spindles with no special treatment or prestress. Each spindle is preloaded with 70% of the $C_a$ given by the manufacturer, where 100% of the $C_a$ is the axial load at which the manufacturer ensures a safe operation of $10^6$ revolutions. In this case, the $C_a$ is chosen with 12kN. With this setup, the camera automatically triggers one camera drive every four hours. Between each image, the spindle is turned by an additional 22.5°, and an area of 150x150 pixels is cropped automatically from the large image.

### D. Dataset for Defect Classification

The dataset is available in (Schlagenhauf & Landwehr et al. 2021) and consists of 21853 150x150 pixel RGB images in the .png format showing areas with and without pitting. The dataset is split such that approximately 50% of the images show pitting. Concretely, the dataset contains 11075 images without pitting and 10778 images with pitting. Each image is assigned with a label $\in \{P, N\}$, where $P$ stands for *pitting* and $N$ stands for *no pitting*. Images followed by an underscore pursue the same logic but are turned by 90° to introduce some variance in the data. This effect can easily be reversed. The authors emphasized the selection of the images regarding the representativeness of the data. The dataset contains all sorts of conditions to which the BSDs are exposed in operation. Figure 5 shows some representative images for the images' whole image set. There are images showing no defect and no pollution like a). There are images showing small pitting with no pollution (b)), small pitting with pollution (c)), no pitting with pollution (d)), and large pitting with (e)) and without (f)) pollution. Hence, the whole spectrum of conditions is covered. Figure 6 and Figure 7 show a larger subset of images with and without pitting. It is obvious that the correct classification of images needs a substantial amount of domain knowledge.

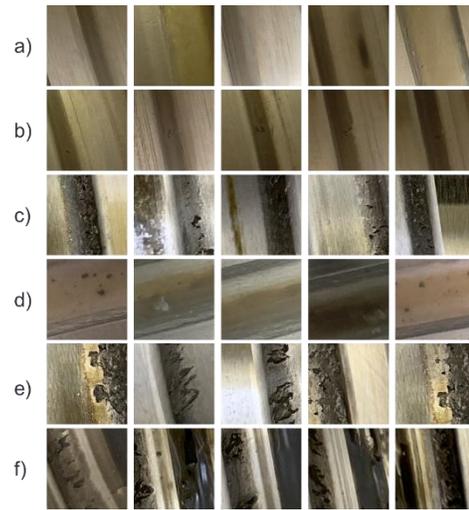

*Figure 5: Subset of the image data taken during the destruction test*

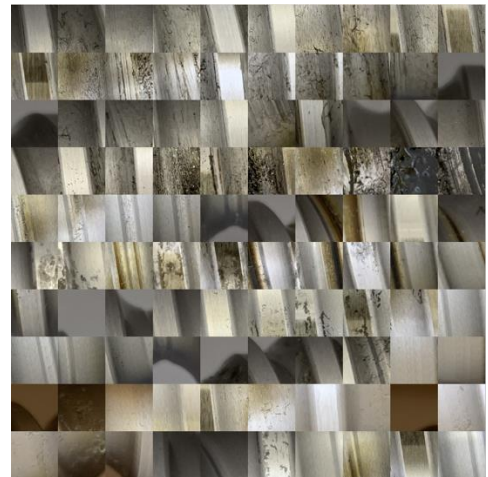

*Figure 6: Subset of images without pitting*

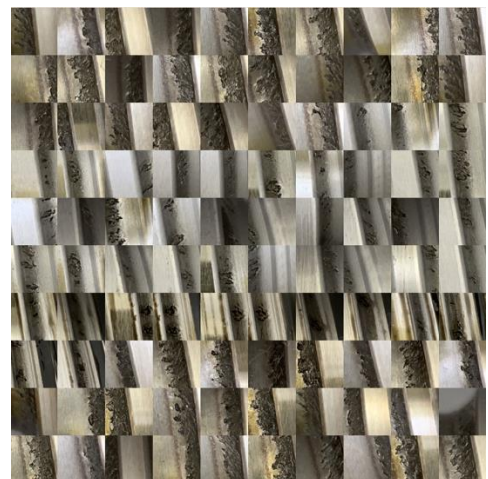

*Figure 7: Subset of images with pitting*

### E. Dataset for Defect Detection/Segmentation

Besides the classification of images, the authors introduce a dataset for instance segmentation which addresses the research problem of image-based size extraction and stands out from the already available datasets for metal surface defect detection like

NEU-DET (Song, Yan 2013), GC10-DET (Lv et al. 2020), or SD-saliency-900 saliency (Song, Song, Yan 2020) with a more suitable representation of real-world problems due to containing a high-class imbalance and pixel-wise annotation masks. Furthermore, this dataset is ideally suited for application areas, namely models that are trained with little data and therefore need to have a high model efficiency.

Condition monitoring enabled by image-based size extraction to detect the current state of a machine tool element, according to (Dong, T. Haftka, H. Kim 2019), can, for example, lead to the reduction of equipment failure cost, improved plant reliability, and optimized maintenance intervals towards a condition-based maintenance strategy and is therefore obviously worthwhile considering. The automatic detection and evaluation of a failure is a critical step towards autonomous production machines.

The introduced dataset is not only for condition-based surface damage detection models on BSDs but also through a size progress detection on image sequences for analysis of wear development over time. This provides the community with a useful dataset for the development and test of wear detection algorithms for all machine tool elements prone to wear which can be recorded by a camera. Three important features are worth noting in particular. The dataset contains tiny damages and hence is suited to develop models especially for the detection of small respectively early defects. In addition to that, the dataset also includes pollution origin from soil which makes detection more difficult together with foreign materials originating from e.g. the production process. As a third feature, the dataset contains the development of the same failures over a period of time. This feature can be used to develop models for the forecasting of failure progressions. To the best of our knowledge, such dataset does not exist in the literature right now. In Figure 8, one exemplary course of an annotated size progress of the dataset is displayed.

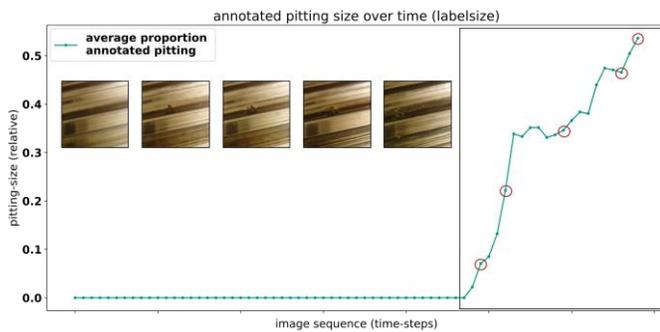

*Figure 8: Annotated pitting size over time of a specific pitting development*

As shown, the graph first remains for approx. 2/3 of the documented time interval at zero due to the fact that there is no surface damage. As soon as pitting occurs, it will only continuously increase its size. The drawn circles represent the size of a single pitting shown in the image cutouts on the left to give an idea about the increasing pitting size. You can also see in the images increasing soiling of the surface and, therefore, there is an increasing difficulty to correctly annotate the pitting. This explains why the shown graph also contains decreasing parts, which is obviously not possible in the real application and opens the possibility to develop models able to cope with this situation.

While classification requires that its data(-points) are assigned to discrete values, such as categories (Cramer, Kamps 2017), and detection can be used for localization of objects within images (Russakovsky et al. 2014), it is recommendable to combine both to detect and classify single objects in images to get as close as possible to the perfect description of an image. Since these dataset annotations can be used for classification as well as detection problems, it is attainable to detect the size of an object and further with the given wear developments forecast the pitting size of the future. Generally, computer vision classification and detection tasks can be divided into four types (Figure 9).

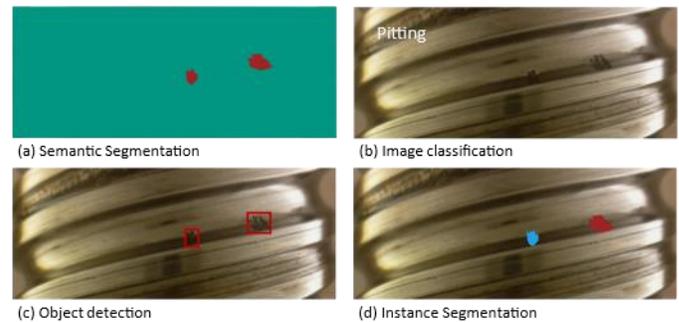

*Figure 9: Different Image classification and Object detection types supported by the dataset*

Instance segmentation (d) as for classification and detection is a pixel-wise object detection method useful for computer vision research tasks like extraction of shape and the exact size of surface damage. Known as one of the most fundamental and challenging tasks in the computer vision research area (Wang et al. 2019), this dataset can also be used for semantic segmentation (a) as a pixel-wise classification with no possibility to distinguish two or more adjacent objects from the same class, an image classification (b) for pitting recognition, and object detection (c) for single object detection.

While most of the related research datasets for damage detection on the metal surface are not annotated for pixel-wise object detection, the introduced dataset cannot only be used for instance segmentation but moreover for the analysis of developments of surface damage over time. The (a) NEU-DET (Song, Yan 2013), shown in Figure 10, for instance, with its 1800 200x200x1 pixel images and six annotation classes (rolled-in scale, patches, crazing, pitted surface, inclusion, scratches) or the (c) GC10-DET (Lv et al. 2020) with its 3570 2048x1000x1 big images and 10 annotation classes (cresent gap, welding line, water spots, silk spot, inclusion, oil spot, crease, punching, waist folding, rolled pit) can only be used for object detection problems.

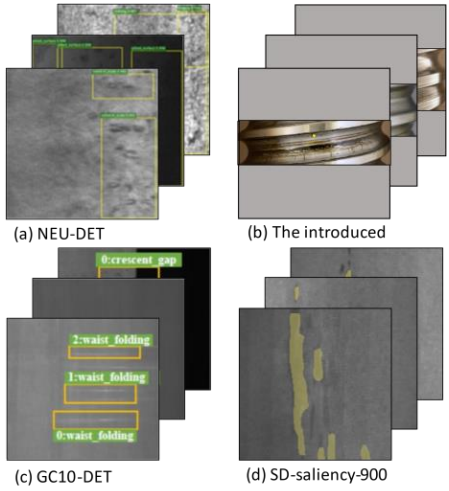

*Figure 10: Different datasets for metal surface damage*

Compared with the instance segmentation (d) SD-saliency-900 dataset (Song, Song, Yan 2020) with its 900 200x200x1 samples, the introduced dataset contains more irrelevant surface information which is an important challenge to address since many real-world problems contain a high-class imbalance (Sammut, Webb 2010).

The dataset contains 1104 channel-3 images with 394 image annotations for the surface damage type "pitting". The annotations made with the annotation tool labelme (Wada 2016) are available in JSON format and hence convertible to VOC and COCO format. All images come from two BSD types. The dataset available for download is divided into two folders, data with all images as JPEG, label with all annotations, and saved_model with a baseline model. The authors also provide a python script to divide the data and labels into three different split types – "train_test_split", which splits images into the same train and test data-split the authors used for the baseline model, "wear_dev_split", which creates all 27 wear developments, and "type_split", which splits the data into the occurring BSD types.

One of the two mentioned BSD types is represented with 69 images and 55 different image sizes. All images with this BSD type come either in a clean or soiled condition.

The other BSD type is shown on 325 images with two image sizes. Since all images of this type have been taken with continuous time, the degree of soiling is evolving.

Also, the dataset contains the above-mentioned 27 pitting development sequences.

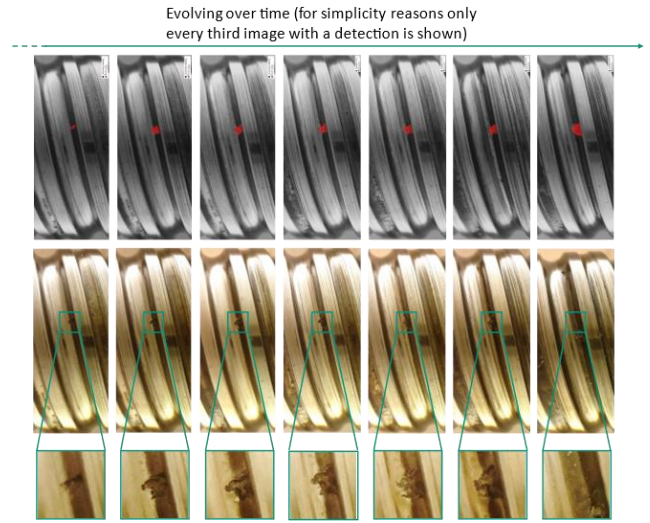

*Figure 11: Pitting process*

Figure 11 shows the evolving pitting development with and without the shown annotations from one of the 27 pitting developments. For convenience, only every third image starting at the beginning of the pitting formation is displayed.

### III. BASELINE MODEL

Regarding the introduced dataset, the authors also present a baseline model. The here used model architecture is a Mask R-CNN (regional Convolutional Network) (He et al. 2017) with an on the COCO dataset (Lin et al. 2014) pretrained Inception ResNet v2 (Szegedy et al. 2016). The Mask R-CNN architecture is composed of two stages, a faster R-CNN with a deep convolutional network composed of Inception v4 and ResNet building blocks united in an Inception ResNet v2 architecture and an FCN (fully convolutional network).

With the chosen architecture, the authors achieved a mIoU (mean intersection over union) baseline score of 0.316. It is noticeable that the model has difficulties predicting small pitting in general (Figure 12 and Figure 13).

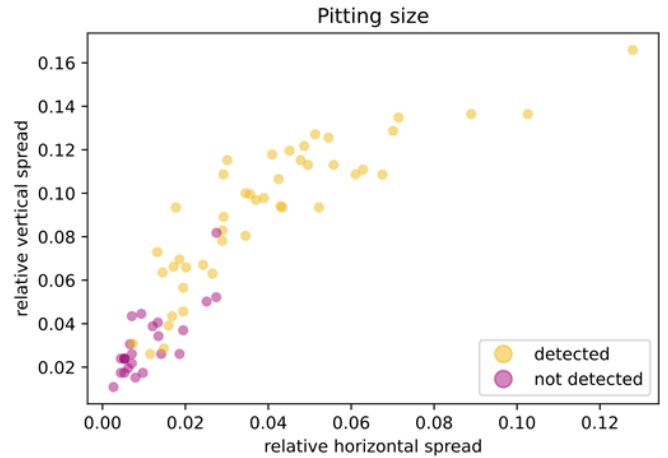

*Figure 12: Relationship between pitting detection and its relative size*

Examining the horizontal and vertical development of pitting and relating it to a binarized model prediction, a zero-one principle - where zero corresponds to "not detected", we can see that pitting detection becomes more reliable as development increases. In Figure 12, the circumstance just described can be readily understood. The relative horizontal spread of the pitting (width) is described on the x-axis and the relative vertical spread (height) is described on the y-axis. The binarization of the detection is represented by the coloring of the points. Figure 13 visualizes the just mentioned circumstance on selected examples.

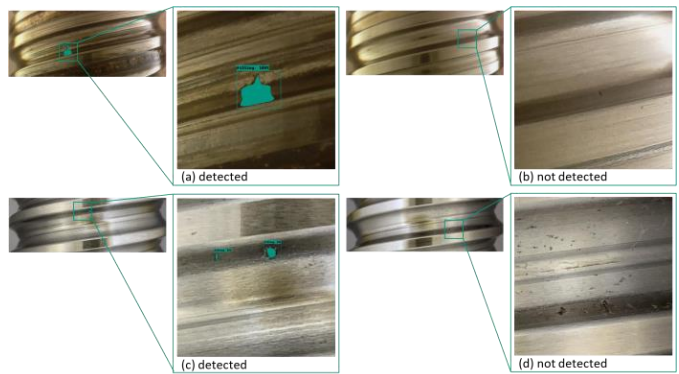

*Figure 13: Prediction examples from the author's model*

The pitting shown in image cutout (a) was due to its large horizontal and vertical spread detected. While the not detected pitting in cutouts (b), (d), and the detected pitting in (c) are relatively small. For convenience, the trained model will be provided.

## IV. CONCLUSION

The need for data is especially prevalent in technical domains where data is limited by the nature of being costly to produce or label. (Fink et al. 2020) To address this issue, real-world technical datasets are necessary to train and test reliable failure detection models. In addition to enabling a machine to automatically predict its remaining useful lifetime, it is necessary to provide it with data depicting different failure stages from which the model can learn. To enable a model to predict the severity of a failure, a huge step towards industrial predictive maintenance and autonomous production machines is done. The dataset presented in this work supports the development of models for this goal.

To the best of our knowledge, the dataset presented in this work is the first technical dataset allowing classification, segmentation, and failure prognostics in one dataset. It allows to develop classical classification models under real-world conditions. Additionally, the dataset can be used to detect tiny defects on polluted surfaces which is an important factor for industrial defect detection applications where it is necessary to detect defects as early as possible. The mentioned possible detection in the latter argument is another feature of the dataset. The third unique feature of the dataset is the fact that the dataset contains the development of defects from early to severe stages. This development can be used to train industrial failure forecasting models.

The authors presented a first ready-to-use baseline model for the detection of surface defects which should serve as a jumping-off point for the development of more powerful model architectures.

The dataset for classification, detection, and forecasting is available in (Schlagenhauf & Landwehr et al. 2021). The trained detection model is available under: https://github.com/2Obe/BSData

This work was supported by the German Research Foundation (DFG) under Grant FL 197/77-1.

## V. REFERENCES

CRAMER, Erhard (Hrsg.); KAMPS, Udo (Hrsg.): *Grundlagen der Wahrscheinlichkeitsrechnung und Statistik : Eine Einführung für Studierende der Informatik, der Ingenieur- und Wirtschaftswissenschaften*. 4., korrigierte und erweiterte Auflage. Berlin : Springer Spektrum, 2017 (Springer-Lehrbuch)

DONG, Ting ; T. HAFTKA, Raphael ; H. KIM, Nam: *Advantages of Condition-Based Maintenance over Scheduled Maintenance Using Structural Health Monitoring System* (2019)

FAGHIH-ROOHI, Shahrzad ; HAJIZADEH, Siamak ; NUNEZ, Alfredo ; BABUSKA, Robert ; SCHUTTER, Bart de: Deep convolutional neural networks for detection of rail surface defects. In: , 2016, S. 2584–2589

FINK, Olga ; WANG, Qin ; SVENSÉN, Markus ; DERSIN, Pierre ; LEE, Wan-Jui ; DUCOFFE, Melanie: *Potential, Challenges and Future Directions for Deep Learning in Prognostics and Health Management Applications*. 2020

FLEISCHER, J. ; BROOS, A. ; SCHOPP, M. ; WIESER, J. ; HENNRICH, H.: *Lifecycle-oriented component selection for machine tools based on multibody simulation and component life prediction*. In: *CIRP Journal of Manufacturing Science and Technology* 1 (2009), Nr. 3, S. 179–184

HABERKERN, Anton: *Leistungsfähigere Kugelgewindetriebe durch Beschichtung*. Universität Karlsruhe, Institut für Werkzeugmaschinen und Betriebstechnik. 1998. URL https://publikationen.bibliothek.kit.edu/27898

HE, Kaiming ; GKIOXARI, Georgia ; DOLLÁR, Piotr ; GIRSHICK, Ross: *Mask R-CNN* (2017)

HE, Ting ; LIU, Ying ; XU, Chengyi ; ZHOU, Xiaolin ; HU, Zhongkang ; FAN, Jianan: *A Fully Convolutional Neural Network for Wood Defect Location and Identification*. In: *IEEE Access* 7 (2019a), S. 123453–123462

HE, Yu ; SONG, Kechen ; MENG, Qinggang ; YAN, Yunhui: *An End-to-end Steel Surface Defect Detection Approach via Fusing Multiple Hierarchical Features*. In: *IEEE Transactions on Instrumentation and Measurement* (2019b), S. 1

HEISE, Wolfgang: *Praxisbuch Zuverlässigkeit und Wartungsfreundlichkeit : R-&-M-Programm für Automobilzulieferer und den Maschinen- und Anlagenbau*. München : Hanser, 2002

KOCH, Christian ; GEORGIEVA, Kristina ; KASIREDDY, Varun ; AKINCI, Burcu ; FIEGUTH, Paul: *A review on computer vision based defect detection and condition assessment of concrete and asphalt civil infrastructure.* In: *Advanced Engineering Informatics* 29 (2015), Nr. 2, S. 196–210

LECUN, Yann ; BENGIO, Yoshua ; HINTON, Geoffrey: *Deep learning.* In: *Nature* 521 (2015), Nr. 7553, S. 436–444

LIN, Tsung-Yi ; MAIRE, Michael ; BELONGIE, Serge ; BOURDEV, Lubomir ; GIRSHICK, Ross ; HAYS, James ; PERONA, Pietro ; RAMANAN, Deva ; ZITNICK, C. Lawrence ; DOLLÁR, Piotr: *Microsoft COCO: Common Objects in Context* (2014)

LV, Xiaoming ; DUAN, Fajie ; JIANG, Jia-Jia ; FU, Xiao ; GAN, Lin: *Deep Metallic Surface Defect Detection: The New Benchmark and Detection Network.* In: *Sensors (Basel, Switzerland)* 20 (2020), Nr. 6

RUSSAKOVSKY, Olga ; DENG, Jia ; SU, Hao ; KRAUSE, Jonathan ; SATHEESH, Sanjeev ; MA, Sean ; HUANG, Zhiheng ; KARPATHY, Andrej ; KHOSLA, Aditya ; BERNSTEIN, Michael ; BERG, Alexander C. ; FEI-FEI, Li: *ImageNet Large Scale Visual Recognition Challenge.* 02.09.2014

SAMMUT, Claude ; WEBB, Geoffrey I.: *Encyclopedia of Machine Learning* (2010)

SCHLAGENHAUF, T.; LANDWEHR, M. & FLEISCHER, J. (2021), *Industrial Machine Tool Element Surface Defect Dataset.* https://publikationen.bibliothek.kit.edu/1000129520

SCHLAGENHAUF, Tobias ; RUPPELT, Peter ; FLEISCHER, Jürgen: *Detektion von frühzeitigen Oberflächenzerrüttungen.* In: *wt Werkstattstechnik online* 110 (2020), 7/8, S. 501–506. URL https://e-paper.vdi-fachmedien.de/webreader-v3/index.html#/2657/50

SCHOPP, Matthias: *Sensorbasierte Zustandsdiagnose und -prognose von Kugelgewindetrieben.* Zugl.: Karlsruhe, Univ., Diss., 2009. Aachen : Shaker, 2009 (Forschungsberichte aus dem wbk, Institut für Produktionstechnik, Karlsruher Institut für Technologie (KIT) 152)

SOMMER, Karl ; HEINZ, Rudolf ; SCHÖFER, Jörg: *Verschleiß metallischer Werkstoffe.* Wiesbaden : Springer Fachmedien Wiesbaden, 2018

SONG, Guorong ; SONG, Kechen ; YAN, Yunhui: *Saliency detection for strip steel surface defects using multiple constraints and improved texture features.* In: *Optics and Lasers in Engineering* 128 (2020), S. 106000

SONG, Kechen ; YAN, Yunhui: *A noise robust method based on completed local binary patterns for hot-rolled steel strip surface defects.* In: *Applied Surface Science* 285 (2013), S. 858–864

SPOHRER, Andreas: *Steigerung der Ressourceneffizienz und Verfügbarkeit von Kugelgewindetrieben durch adaptive Schmierung.* Dissertation. 1. Auflage. Düren : Shaker, 2019 (Forschungsberichte aus dem wbk, Institut für Produktionstechnik Universität Karlsruhe 225)

SZEGEDY, Christian ; IOFFE, Sergey ; VANHOUCKE, Vincent ; ALEMI, Alex: *Inception-v4, Inception-ResNet and the Impact of Residual Connections on Learning* (2016)

TANG, Tien T. ; ZAWASKI, Janice A. ; FRANCIS, Kathleen N. ; QUTUB, Amina A. ; GABER, M. Waleed: *Image-based Classification of Tumor Type and Growth Rate using Machine Learning: a preclinical study.* In: *Scientific reports* 9 (2019), Nr. 1, S. 12529

WADA, Kentaro: *labelme: Image Polygonal Annotation with Python* (2016). URL https://github.com/wkentaro/labelme – Review date 08.02.2021

WANG, Xinlong ; KONG, Tao ; SHEN, Chunhua ; JIANG, Yuning ; LI, Lei: *SOLO: Segmenting Objects by Locations* (2019)